%% file: main.tex
\documentclass[10pt,twocolumn,letterpaper]{article}
\usepackage[accsupp]{axessibility}
\usepackage{cvpr}              % To produce the CAMERA-READY version
\usepackage{afterpage}
\input{preamble}

\input{macro}

\definecolor{cvprblue}{rgb}{0.21,0.49,0.74}
\usepackage[pagebackref,breaklinks,colorlinks,citecolor=cvprblue]{hyperref}

\def\confName{CVPR}
\def\confYear{2024}

\title{Driving Everywhere with Large Language Model Policy Adaptation}

\author{Boyi Li$^1$ 
\ 
Yue Wang$^{1,2}$
\
Jiageng Mao$^2$
\
Boris Ivanovic$^1$
\
Sushant Veer$^1$
\
Karen Leung$^{1,3}$ 
\
Marco Pavone$^{1,4}$ \\
$^1$NVIDIA \quad $^2$ University of Southern California \quad $^3$ University of Washington \quad $^4$ Stanford University
}

\begin{document}
\maketitle
\input{sec/abstract}    
\input{sec/intro}
\input{sec/relatedworks}
\input{sec/method}

\input{sec/LLaDA_for_AV}
\input{sec/experiments}

\input{sec/conclusion}

{
    \small
    \bibliographystyle{ieeenat_fullname}
    \bibliography{reference}
}

\end{document}

%% file: preamble.tex
\usepackage[dvipsnames]{xcolor}

\newcommand{\greentext}[1]{\textcolor{green}{#1}}
\newcommand{\redtext}[1]{\textcolor{red}{#1}}
\newcommand{\purpletext}[1]{\textcolor[rgb]{0.298,0,0.6}{#1}}
\usepackage{graphicx}
\usepackage{booktabs}
\usepackage{multicol}
\usepackage{multirow}
\usepackage{colortbl}

%% file: macro.tex
\newcommand{\method}{LLaDA} 

\newcommand{\module}{Traffic Rule Extractor}
\newcommand{\moduleshort}{TRE} 

\definecolor{bleudefrance}{rgb}{0.19, 0.55, 0.91}
\definecolor{greenncs}{rgb}{0.0, 0.62, 0.42}
\definecolor{greenpigment}{rgb}{0.0, 0.65, 0.31}
\definecolor{jazzberryjam}{rgb}{0.65, 0.04, 0.37}

%% file: sec/abstract.tex
\begin{abstract}
Adapting driving behavior to new environments, customs, and laws 
is a long-standing problem in autonomous driving, precluding the widespread deployment of autonomous vehicles (AVs). In this paper, we present LLaDA, a simple yet powerful tool that enables human drivers and autonomous vehicles alike to drive everywhere by adapting their tasks and motion plans to traffic rules in new locations. LLaDA achieves this by leveraging the impressive zero-shot generalizability of large language models (LLMs) in interpreting the traffic rules in the local driver handbook.
Through an extensive user study, we show that LLaDA's instructions are useful in disambiguating in-the-wild unexpected situations. We also demonstrate LLaDA's ability to adapt AV motion planning policies in real-world datasets; LLaDA outperforms baseline planning approaches on all our metrics. 
Please check our website for more details: \href{https://boyiliee.github.io/llada/}{LLaDA}.

\end{abstract}

%% file: sec/intro.tex
\section{Introduction}
\label{sec:intro}

Despite the rapid pace of progress in autonomous driving,
autonomous vehicles (AVs) continue to operate primarily in geo-fenced areas. A key inhibitor for AVs to be able to drive everywhere is the variation in traffic rules and norms across different geographical regions. Traffic rule differences in different geographic locations can range from significant (e.g., left-hand driving in the UK and right-hand driving in the US) to subtle (e.g., right turn on red is acceptable in San Francisco but not in New York city). In fact, adapting to new driving rules and customs is difficult for humans and AVs alike; failure to adapt to local driving norms can lead to unpredictable and unexpected behaviors which may result in unsafe situations \cite{koppenborg2017human,zhang2017plan}. Studies have shown that tourists are more susceptible to accidents \citep{psarras2023covid,rossello2011road} that can sometimes result in injury or death \citep{laHondaAccident}. This calls for a complete study of policy adaption in current AV systems.

\input{imgs/teaser1/figure}

At the same time, LLMs have recently emerged as front-runners for zero- or few-shot adaptation to out-of-domain data in various fields, including vision and robotics~\citep{codeaspolicies2022,li2023itp,elhafsi2023semantic}. 
Inspired by these works, our goal is to build a \textbf{L}arge \textbf{La}nguage \textbf{D}riving \textbf{A}ssistant (\method{}) that can rapidly adapt to local traffic rules and customs (\autoref{fig:teaser}). Our method consists of three steps: First, we leverage existing methods to generate an executable policy; second, when presented with an unexpected situation described in natural language (either by a human prompt or a VLM such as GPT-4V \cite{openai2023gpt4} or LINGO-1 \cite{lingo1}), we leverage a \module{} (\moduleshort{}) to extract informative traffic rules relevant to the current scenario from the local traffic code; finally, we pass the \moduleshort{}'s output along with the original plan to a pre-trained LLM (GPT-4V \cite{openai2023gpt4} in this paper) to adapt the plan accordingly. We test our method on the nuScenes \cite{caesar2020nuscenes} dataset and achieve improvements in motion planning under novel scenarios. We also provide extensive ablation studies and visualizations to further analyze our method.

\textbf{Contributions.} Our core contributions are three-fold: 
\begin{enumerate}
    \item We propose \method{}, a training-free mechanism to assist human drivers and adapt autonomous driving policies to new environments by distilling leveraging the zero-shot generalizability of LLMs. 
    \item \method{} can be immediately applied to any autonomous driving stack to improve their performance in new locations with different traffic rules.
    \item Our method achieves performance improvements over previous state-of-the-arts, as verified by user studies and experiments on the nuScenes dataset. 
\end{enumerate}

%% file: imgs/teaser1/figure.tex
\begin{figure}
\centering
\includegraphics[width=1.0\linewidth]{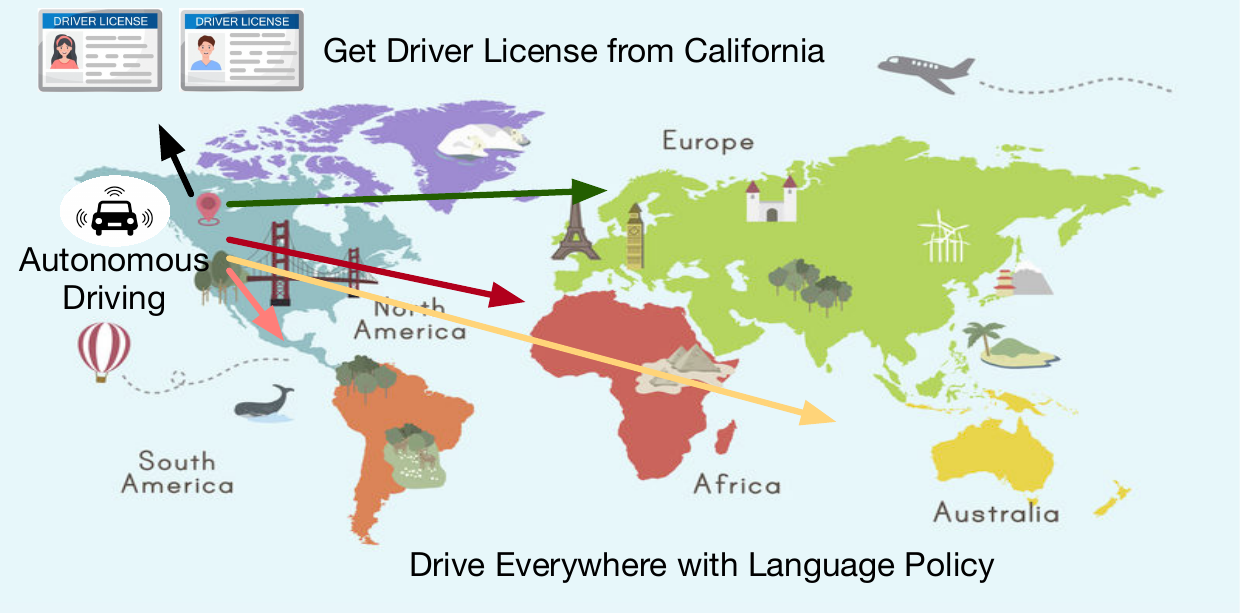}
\caption{\method{} enables drivers to obtain instructions in any region all over the world. For instance, the driver gets a driver's license in California, USA, our system enables providing prompt instructions when the driver drives in different regions with different situations.}
\label{fig:teaser}
\end{figure}

%% file: sec/relatedworks.tex
\section{Related Works}
\label{sec:relatedworks}

\textbf{Traffic Rules in AV Planning.} Researchers have explored the possibility of embedding traffic rules in the form of metric temporal logic (MTL) formulae \cite{manas2022legal}, linear temporal logic (LTL) formulae \cite{esterle2020formalizing,maierhofer2020formalization,karlsson2020intention}, and signal temporal logic (STL) formulae \cite{veer2022receding,xiao2021rule}. Expressing the entire traffic law as logic formulae is not scalable due to the sheer number of rules and the exceptions that can arise. Furthermore, adapting to traffic rules in a new region still requires the cumbersome encoding of new traffic rules in a machine-readable format. This challenge was highlighted in \cite{manas2022robust}, where the use of a natural language description of traffic rules was proposed as a potential solution. There is a dearth of literature on directly using the traffic rule handbook in its natural language form for driving adaptation to new locations, and it is precisely what we aim to achieve in this paper.

\textbf{LLMs for Robotic Reasoning.} Recently, many works have adopted LLMs to tackle task planning in robotics. These methods generally leverage LLMs' zero-shot generalization and reasoning ability to design a feasible plan for robots to execute. Of note, PaLM-E~\cite{driess2023palme} develops an embodied multi-modal language model to solve a broad range of tasks including robotic planning, visual question answering, and captioning; this large model serves as a foundation for robotic tasks. VLP~\cite{du2023video} further enables visual planning for complex long-horizon tasks by pretraining on internet-scale videos and images. Code-As-Policies~\cite{codeaspolicies2022} re-purposes a code-writing LLM to generate robot policy code given natural language commands; it formulates task planning as an in-context code generation and function call problem. ITP~\cite{li2023itp} further proposes a simple framework to perform interactive task planning with language models, improving upon Code-As-Policies. Inspired by these works, our method also leverages LLMs for autonomous driving. However, the key difference is that our method focuses on policy adaptation via LLMs rather than the wholesale replacement of modules with LLMs.  

\textbf{LLMs for Autonomous Driving.} Most autonomous driving pipelines consist of perception, prediction, planning, and control, which have been significantly advanced by machine learning and deep neural networks in recent years.
Despite such tremendous progress, both perception and planning are generally non-adaptive, preventing AVs from generalizing to any in-the-wild domain. Recent works leverage foundation models to provide autonomous driving pipelines with common sense reasoning ability. \citet{wang2023drive} proposes a method to extract nuanced spatial (pixel/patch-aligned) features from Transformers to enable the encapsulation of both spatial and semantic features. GPT-Driver~\cite{mao2023gptdriver} finetunes GPT-3.5 to enable motion planning and provide chain-of-thought reasoning for autonomous driving. DriveGPT4~\cite{xu2023drivegpt4} further formulates driving as an end-to-end visual question answering problem. Most recently, MotionLM~\cite{seff2023motionlm} represents continuous trajectories as sequences of discrete motion tokens and casts multi-agent motion prediction as a language modeling task over this domain. Our work also leverages LLMs for policy adaption, however, we do not fine-tune or train a new foundation model. Instead, our method capitalizes on GPT-4 to perform direct in-context reasoning. 

In parallel, there has been a plethora of literature on AV out-of-domain (OoD) generalization and detection \cite{filos2020can,itkina2023interpretable,veer2023multi,amini2022vista,hendrycks2019scaling,farid22arxiv-predictionAnomaly,mcallister2019robustness,IvanovicHarrisonEtAl2023}. However, the vast majority of such works focus on low-level tasks (e.g., transferring perception models to data from different sensor configurations \cite{amini2022vista}, adapting prediction methods to behaviors from different regions~\cite{IvanovicHarrisonEtAl2023}, etc.) and less on higher-level semantic generalization~\cite{elhafsi2023semantic}, which our work focuses on.

%% file: sec/method.tex
\section{Driving Everywhere with Large Language Model Policy Adaptation}
\label{sec:method}

In this section, we will introduce our method, \method{}, for adapting motion plans to traffic rules in new geographical areas and discuss all its building blocks.

\input{imgs/main/figure}
\method{} receives four inputs, all in the form of natural language: (i) a nominal execution plan, (ii) the traffic code of the current location, (iii) a description of the current scene from the ego's perspective, and (iv) a description of any ``unexpected" scenario that may be unfolding. \method{} ingests these four inputs and outputs a motion plan -- also represented in natural language -- that addresses the scenario by leveraging the local traffic rules. The nominal execution plan can be generated by a human driver.
Similarly, the scene description and the unexpected scenario description can be generated by a human or a VLM. The unique traffic code in the current location is the text of the entire driver handbook that describes the rules of the road for that location. 
Under normal circumstances, the unexpected situation input defaults to \texttt{normal status}; however, if something out-of-the-ordinary unfolds, such as the ego vehicle getting honked or flashed at, or if the ego driver notices something unusual in the environment (e.g., an animal on the road), the appropriate text description of the scenario can be supplied to \method{}. To make the role of \method{} more concrete, consider an example: An AV is operating in New York City (NYC) and the nominal motion plan for the vehicle is to \texttt{turn right} at a \texttt{signalized intersection with a red light}. The AV was \texttt{honked at by cross traffic} which is unexpected. \method{} will take these inputs along with NYC's driver manual and adapt the motion plan to \texttt{no right turn on a red light} because NYC traffic law prohibits right turns on red~\cite{NYC_RTOR}. In the remainder of this section, we will discuss the building blocks of \method{}, illustrated in \autoref{fig:main}.

\textbf{\module{}.} 
\input{imgs/tre/figure}
Passing the entire driver handbook to the LLM is superfluous as we only need the traffic rules relevant to the current scenario that the vehicle finds itself in. In fact, extraneous information of the traffic code can hurt the LLM Planner's performance. To achieve this \emph{task-relevant} traffic rule extraction we use the \module{} (\moduleshort{}). \moduleshort{} uses the nominal execution plan and the description of the unexpected scenario to extract keywords in the traffic code of the current location, which are further used to extract paragraphs that comprise these keywords. We use \module{} (\moduleshort{})
to identify the most relevant keywords and paragraph extraction; see \autoref{fig:tre} for an illustration of \moduleshort{}'s operation. We could observe that \moduleshort{} is simple yet efficient in extracting key paragraphs from the unique traffic code, it first generates a prompt and finds keywords in the organized prompt using GPT-4. Then we find the keywords in the unique traffic code in the current location. By organizing the processed guidelines and prompt, we can obtain a new plan accurately by using GPT-4 twice. After obtaining the relevant paragraph, we input the organized information from \moduleshort{} into an LLM (GPT-4) to obtain the final new plan, referred to as the LLM Planner.

%% file: imgs/main/figure.tex
\begin{figure*}
\centering
\includegraphics[width=1.0\linewidth]{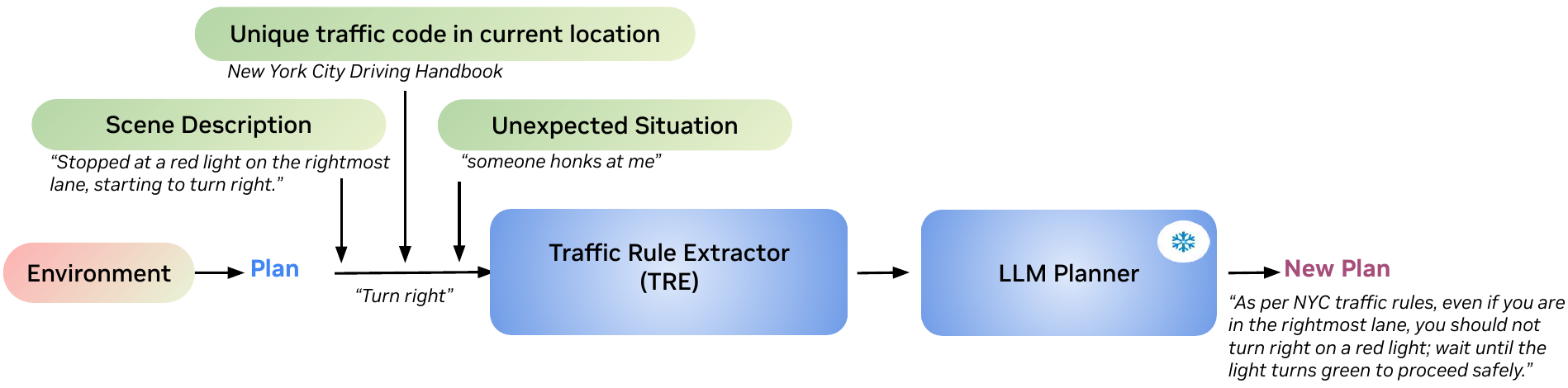}
\caption{Overview of \method{}. In this illustration, the driver learned how to drive in California but now needs to drive in New York City. However, the road situation, traffic code, and unexpected situations are different. In our system, we consider three inputs: initial plan (``\textit{Turn right}"), unique traffic code in current location (\textit{New York City Driving Handbook}), and unexpected situation (``\textit{someone honks at me}"). We will feed these three inputs into a \module{} (\moduleshort{}), which aims to organize and filter the inputs and feed the output into the frozen LLMs to obtain the final new plan. In this paper, we set GPT-4 as our default LLM. }
\label{fig:main}
\end{figure*}

%% file: imgs/tre/figure.tex
\begin{figure*}
\centering
\includegraphics[width=1.0\linewidth]{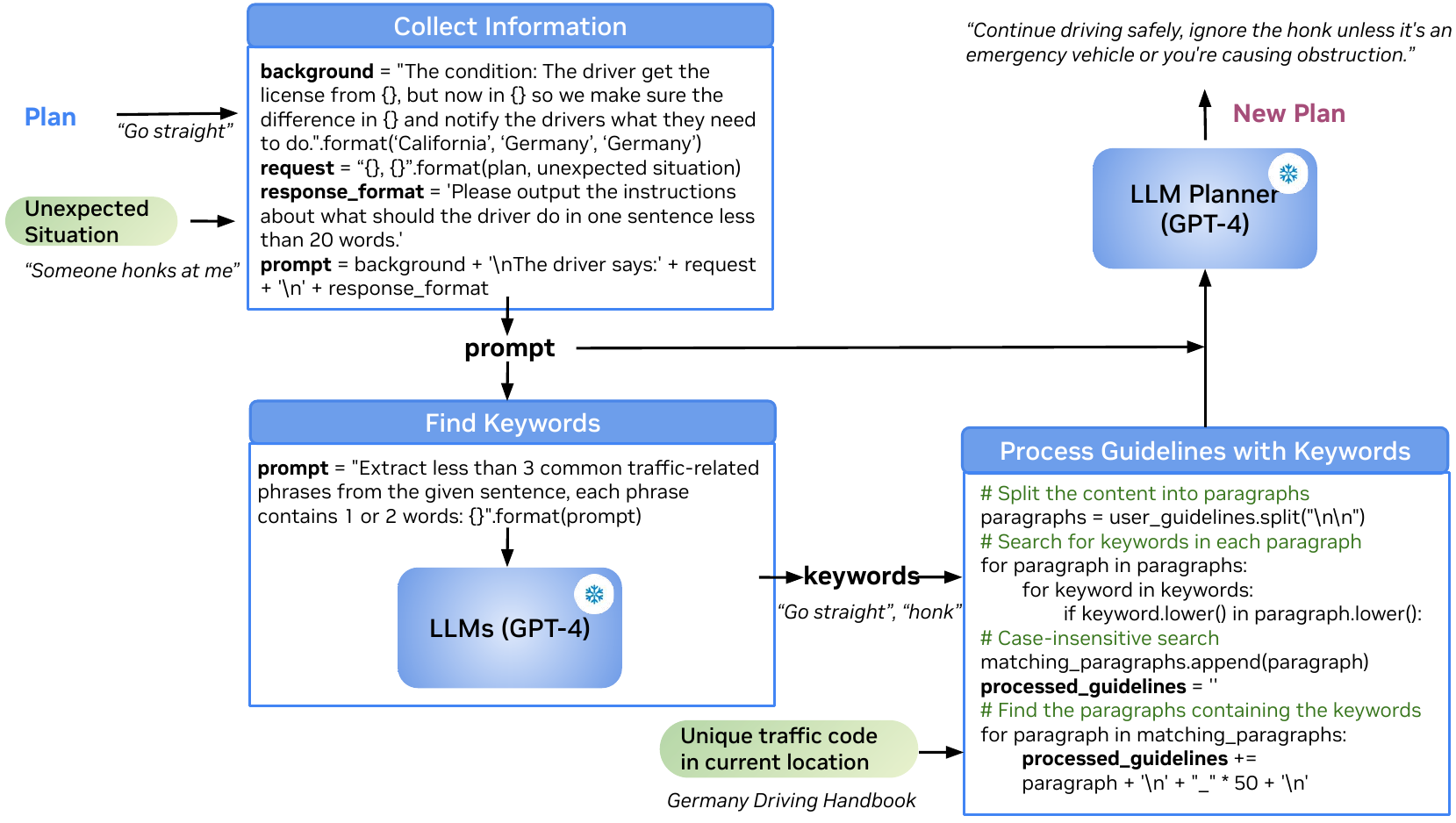}
\caption{Details of \module{} (\moduleshort{}). As is shown in the figure, we first organize the information (such as locations, ``Turn right" and ``someone honks at me" ) into a prompt. Then we feed the prompt to find the one or two keywords using GPT-4. To guarantee the search quality, each keyword contains one or two words. Then we find the key paragraphs that contain extracted keywords in the unique traffic code. In this way, we could filter out the necessary information and only organize the valuable material into GPT-4 to obtain the final new plan.}
\label{fig:tre}
\end{figure*}

%% file: sec/LLaDA_for_AV.tex
\section{Applications of \method{}}

\method{} is a general purpose tool for seamlessly adapting driving policies to traffic rules in novel locations. We see two main applications that can benefit from \method{}:

\textbf{Traffic Rule Assistance for Tourists.} Standalone, \method{} can serve as a guide for human drivers in new locations. We envision an interface wherein a human driver, when encountered with an unexpected scenario, can query \method{} in natural language via a speech-to-text module on how to resolve it. As described in Section~\ref{sec:method}, \method{} can take this natural language description of the scene, the unexpected scenario, and the nominal execution plan and provide a new plan which adheres to the local traffic laws. It is worth pointing out that in its current form, \method{} cannot provide plan corrections unless queried by the human driver. This limits its usability to scenarios where the human driver becomes aware that they are in an unexpected scenario. Extending \method{} to automatically provide plan corrections requires the development of an \emph{unexpected scenario detector and translator}, which is beyond the scope of this current work and will be explored as part of our future work. We conducted a survey to garner human feedback about the usefulness and accuracy of \method{} in some challenging traffic rule scenarios -- the results are discussed in Section~\ref{sec:experiments}.

\input{imgs/gpt_driver/figure}
\textbf{AV Motion Plan Adaptation.}
\label{subsec:adapt-AV-plan}We can also leverage \method{}'s traffic law adaptation ability in an AV planning stack to automatically adapt AV plans to the rules of a new geographical location. This can be achieved by interfacing \method{} with any motion planner capable of generating high-level semantic descriptions of its motion plan (e.g., GPT-driver \cite{mao2023gptdriver}) and a VLM (e.g., GPT-4V) that can translate the scene and the unexpected scenario into their respective textual descriptions. \method{} then adapts the nominal execution plan and communicates it to a downstream planner that updates the low-level waypoint trajectory for the AV. Our approach for using \method{} to adapt AV motion plans is summarized in \autoref{fig:gpt_driver}. We demonstrate the benefits that \method{} can deliver to AV planning in our experiments where a nominal planner trained in Singapore is deployed in Boston; more details regarding the experiments are provided in Section~\ref{sec:experiments}.

%% file: imgs/gpt_driver/figure.tex
\begin{figure*}
\centering
\includegraphics[width=1.0\linewidth]{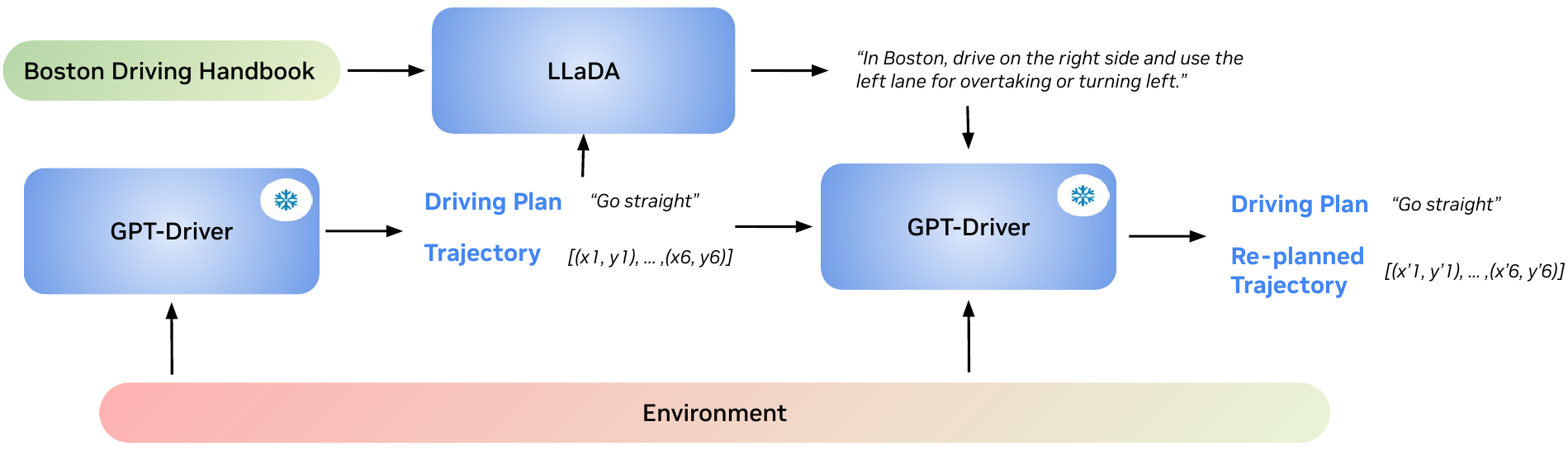}
\caption{Combining \method{} with GPT-Driver for motion planning on the nuScenes dataset.}
\label{fig:gpt_driver}
\end{figure*}

%% file: sec/experiments.tex
\section{Experiments}
\label{sec:experiments}

\subsection{Implementation Details}
Since \method{} takes advantage of large pre-trained language models, our method is training-free and easily be applied to any existing driving system. \method{} could be run with a single CPU. In this paper, we assume the driver obtains driver's license from California as the default setting.

\subsection{\method{} Examples}
We show a full set of functions of \method{} in Figure~\ref{fig:example1}. \method{} enables the system to provide the most updated instructions based on local traffic rules, we show the basic functions of \method{} and display how it works when the drivers are in different places, in diverse unexpected situations, or with diverse plans under various environments. We could observe that \method{} is robust to distinct conditions. We also notice that without the driving handbook, the model cannot provide accurate information. We assume this is because GPT-4 may not be able to provide detailed instructions without the context or complex prompt tuning, while \method{} could successfully alleviate this problem and generate reasonable instructions with emphasis on the specific local traffic rule and driver's request.

\input{imgs/example1/figure}

\subsection{Inference on Random Nuscenes/Nuplan Videos}
Nuscenes~\citep{caesar2020nuscenes} and Nuplan~\citep{caesar2021nuplan} datasets are two of the most used dataset for autonomous driving. Nuscenes is the first dataset to carry the full autonomous vehicle sensor suite and Nuplan is the world's first closed-loop ML-based planning benchmark for autonomous driving. However, Nuscenes only contains 12 simple instructions such as ``accelerate" and Nuplan only has 74 instructions (scenario types) such as ``accelerating at stop sign no crosswalk", which may not provide constructive and efficient instructions for drivers in different locations. \method{} could successfully address this problem and can be applied to random videos. We first show Nuscenes example in \autoref{fig:exp_nuscenes}. We also show Nuplan examples in \autoref{fig:exp_nuplan}. It is obvious that \method{} works for random videos under diverse scenarios, achieving driving \textbf{everywhere} with language policy. 
\input{imgs/exp_nuscenes/figure}
\input{imgs/exp_nuplan/figure}

\subsection{Challenging Situations}
To further verify the efficiency of \method{}, we consider several challenging cases and compare the results with and without our approach. Also, since GPT-4 could translate different languages at the meanwhile, \method{} is able to process different language inputs and output the corresponding instructions (See row 5). We display the results in \autoref{tab:challenging}. In example 1, in NYC there is no right-turn on red, which is allowed in San Francisco. In example 2, \method{} can point out something relating to Rettungsgasse (move to the right). This is because in the US the rule is that everyone pulls over to the right, but this is not standard in Germany. In example 3, \method{} is able to point out that we should overtake the slow car safely from the right lane, since overtaking on the left is illegal in London. For example 4, our system could point out that an unprotected right in England (left-driving system) requires checking the traffic coming at you as you will have to cut through upcoming traffic. Both should mention checking for pedestrians. For example 5, since the driver is on the Autobahn, where being in the left lane typically requires driving at very high speeds compared to American or many other countries' highway speed limits. On the German Autobahn, the absence of a speed limit means that drivers instead adopt different "speed zones" per lane, with the leftmost being the fastest and the rightmost being the slowest. For example 6, Amish communities exist in the US and Canada (primarily in the northeast USA), and they frequently have horse-pulled carriages on roads. So our system successfully provides the instructions to give right-of-way to horses.

\input{tables/challenging}

\subsection{Evaluator-based Assessment.}
We conducted an evaluator-based assessment to further validate the usefulness of videos generated with \method{}. Here we show the corresponding questionnaire, we list the questions in \href{https://forms.gle/NRjbNwPSZh5ZbzxdA}{this Google Form}. 
We provided location, scenario, unexpected situation, relevant local law as conditions, and \method{} output as driving assistant instructions. Here is the relevant local law that indicates what we want to pay attention to while driving. We asked two questions to 24 participants about each of the 8 cases. $54.2\%$ participants have more than 10 years of driving experience and $20.8\%$ participants have 5-10 years driving experience. Also, $75\%$ participants are from the United States. In our assessment, we ask two questions: ``\textit{Does the instruction follow the relevant local law?}'' and ``\textit{How useful is the instruction?}''. The results show that $70.3\%$ participants think the instructions strictly follow the relevant local law, and $82.8\%$ participants find the instructions are very or extremely helpful for them. This highlights that \method{} brings about a significant enhancement in the performance of baseline video diffusion models in both the alignment and visual quality. 
\input{imgs/viz/figure}
\subsection{Comparison on Motion Planning}
We conduct experiments on the nuScenes dataset to validate the effectiveness of \method{} in motion planning. NuScenes consists of perception and trajectory data collected from Singapore and Boston, which have different traffic rules (e.g., driving side difference). Specifically, we first utilize GPT-Driver~\cite{mao2023gptdriver} to generate an initial driving trajectory for a particular driving scenario, and then we leverage \method{} to generate guidelines for GPT-Driver to re-generate a new planned driving trajectory. Since \method{} provides country-specific guidelines, we fine-tuned the GPT-Driver on the \textit{Singapore} subset of the nuScenes dataset and evaluated the performances of GPT-Driver and \method{} on the \textit{Boston} subset of the nuScenes validation set. We follow~\cite{hu2022stp3, mao2023gptdriver} and leverage L2 error (in meters) and collision rate (in percentage) as evaluation metrics. The average L2 error is computed by measuring each waypoint's distance in the planned and ground-truth trajectories. It reflects the proximity of a planned trajectory to a human driving trajectory. The collision rate is computed by placing an ego-vehicle box on each waypoint of the planned trajectory and then checking for collisions with the ground truth bounding boxes of other objects. It reflects the safety of a planned trajectory. We follow the common practice in previous works and evaluate the motion planning result in the $3$-second time horizon. \autoref{tab:plan} shows the motion planning results. With the guidelines provided by \method{}, GPT-Driver could adapt its motion planning capability from Singapore to Boston and reduce planning errors.

\input{tables/planning}

\subsection{Ablation Study on Potential Safety Issues}
There might be concerns that since \method{} is based on LLMs, it might generate prompts that might provide dangerous instructions. To alleviate this concern, we evaluate \method{} on three different critical cases with diverse countries and unexpected situations: 1) when facing a stop sign, whether \method{} suggests ``stop" or not. 2) When facing the red light, whether \method{} suggests ``stop"/specific safe notifications or not. 3) When it rains heavily, whether \method{} suggests ``slow down" or not. 4) When a pedestrian walks across the street, whether \method{} suggests to ``yield to the pedestrian" or not. For each case, we evaluate 50 random examples and report the average score of each case. For each example, if the answer will cause potentially dangerous behavior, we will treat it as an error. We observe that \method{} achieves $0\%$ error rate for all 4 cases. In case ``stop sign", where all the instructions suggest ``Wait at the stop sign until it's safe". In case ``red light", where all the instructions suggest ``come to a complete stop at the red light" or "wait until it turns green". In case ``it rains heavily", where all the instructions suggest ``turn on the headlight" and "reduce speed". In case ``A pedestrian walks across the street", where all the instructions suggest ``yield to the pedestrian". We didn't notice any potentially harmful instructions that might cause dangerous behavior,  which ensures user safety by directing them toward appropriate behavior.

\subsection{Combining with GPT-4V}
Our approach can be combined with different systems to improve the functionality to provide accurate information for humans. In this section, we study the case by combining \method{} with GPT-4V. GPT-4V is the vision branch of GPT-4 which can output corresponding captions to describe the scenes based on a visual input. We randomly pick two scenes from Youtube~\footnote{Source:\url{https://www.youtube.com/watch?v=Boh66Pjjiq0} and \url{https://www.youtube.com/watch?v=xn_mSqTrOUo}} and ask GPT-4V to provide additional captions, we add these additional captions to the user's request. We show an example in \autoref{fig:gpt4v}, it could be observed that \method{} could process the information GPT-4V very well and provide accurate instructions based on given captions. 
\input{imgs/exp_youtube/figure}

%% file: imgs/example1/figure.tex
\begin{figure*}
\centering
\includegraphics[width=1.0\linewidth]{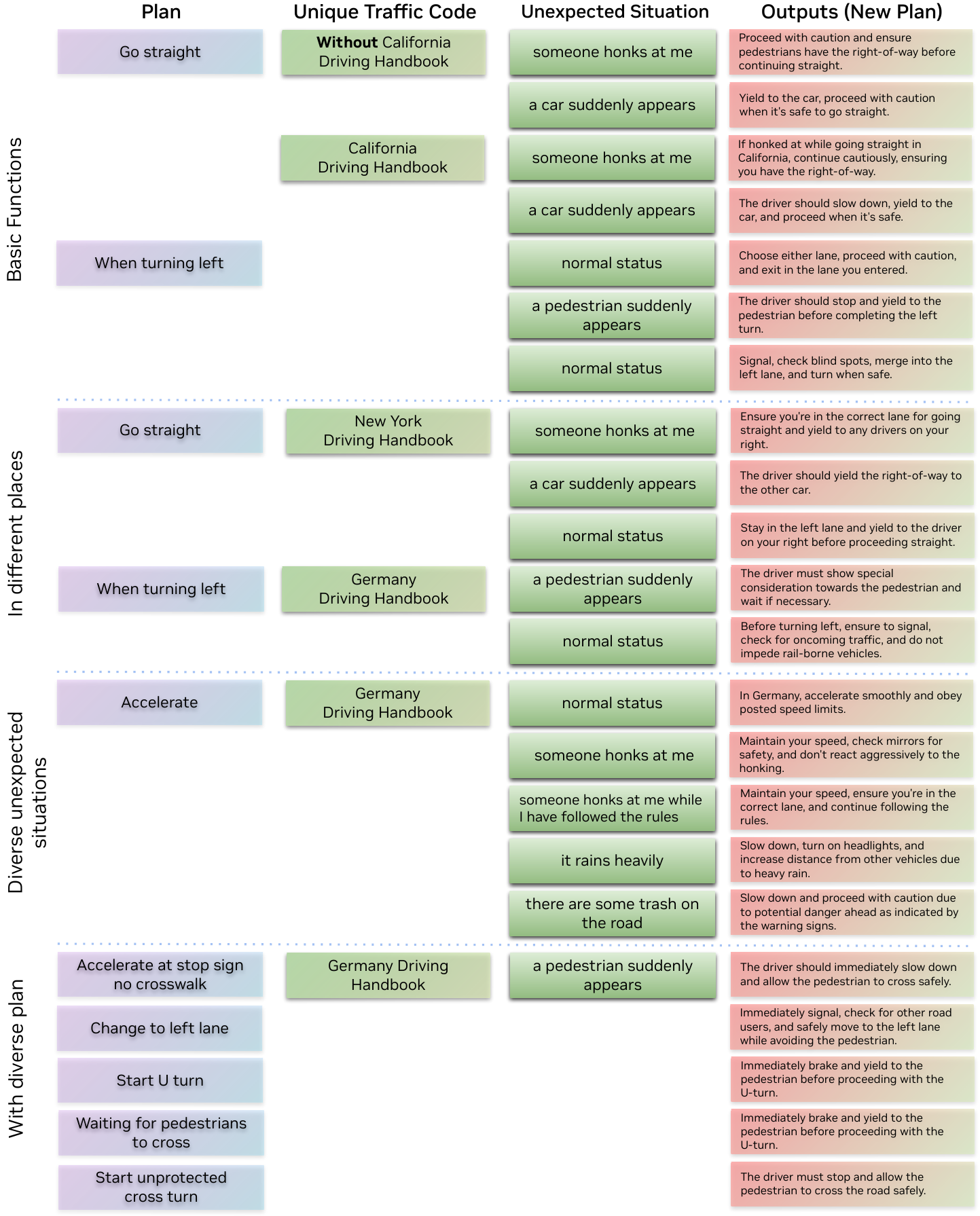}
\caption{We show a few examples of \method{} to help drivers drive everywhere with language policy. We show \method{} could help the drivers obtain prompt notification and correct their corresponding behaviors in different countries with diverse plans and diverse unexpected situations. Also, it is obvious that LLM cannot provide accurate instruction based on each location without the background of the traffic code.}
\label{fig:example1}
\end{figure*}

%% file: imgs/exp_nuscenes/figure.tex
\begin{figure*}
\centering
\includegraphics[width=1.0\linewidth]{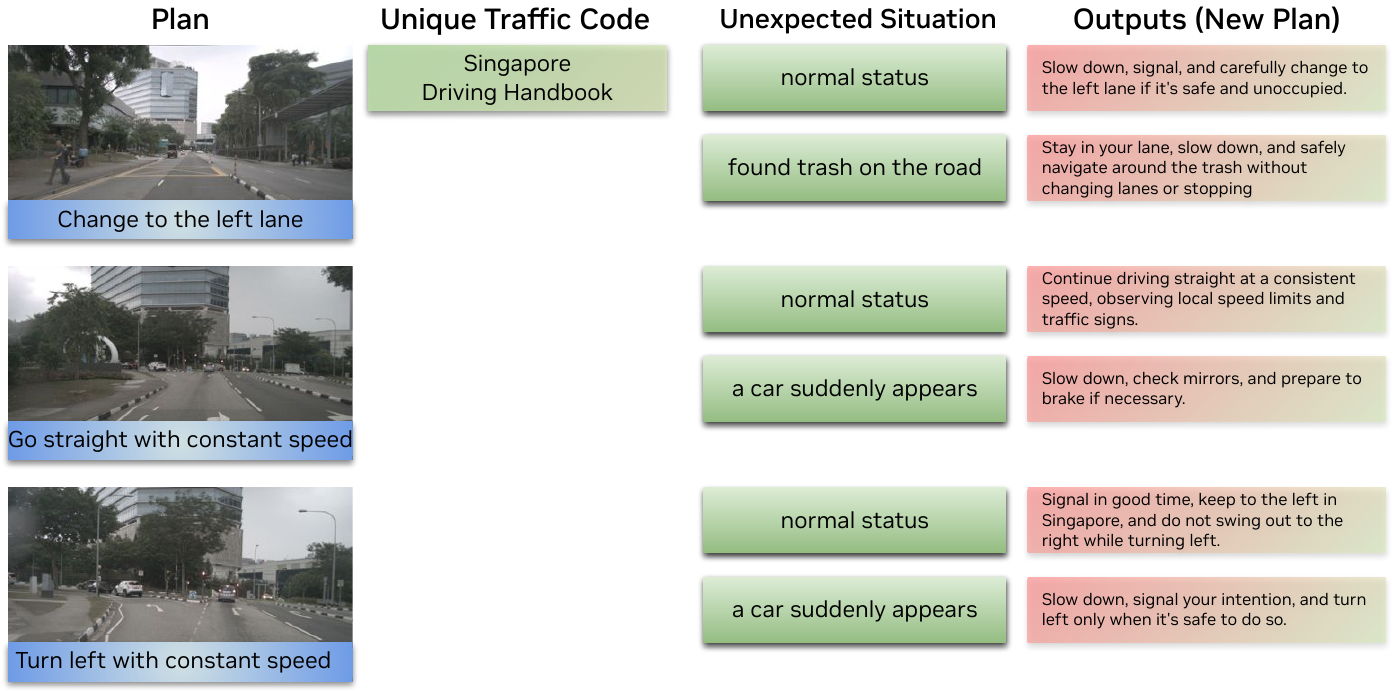}
\caption{Inference on a random Nuscenes video in Singapore. Since different from most countries, Singapore has a right-hand traffic system, which requires distinct behavior in comparison with California's (left-hand traffic system). For example, when ``Turn left with constant speed" and ``normal status", \method{} suggest the driver to ``keep to the left in Singapore".}
\label{fig:exp_nuscenes}
\end{figure*}

%% file: imgs/exp_nuplan/figure.tex
\begin{figure*}
\centering
\includegraphics[width=1.0\linewidth]{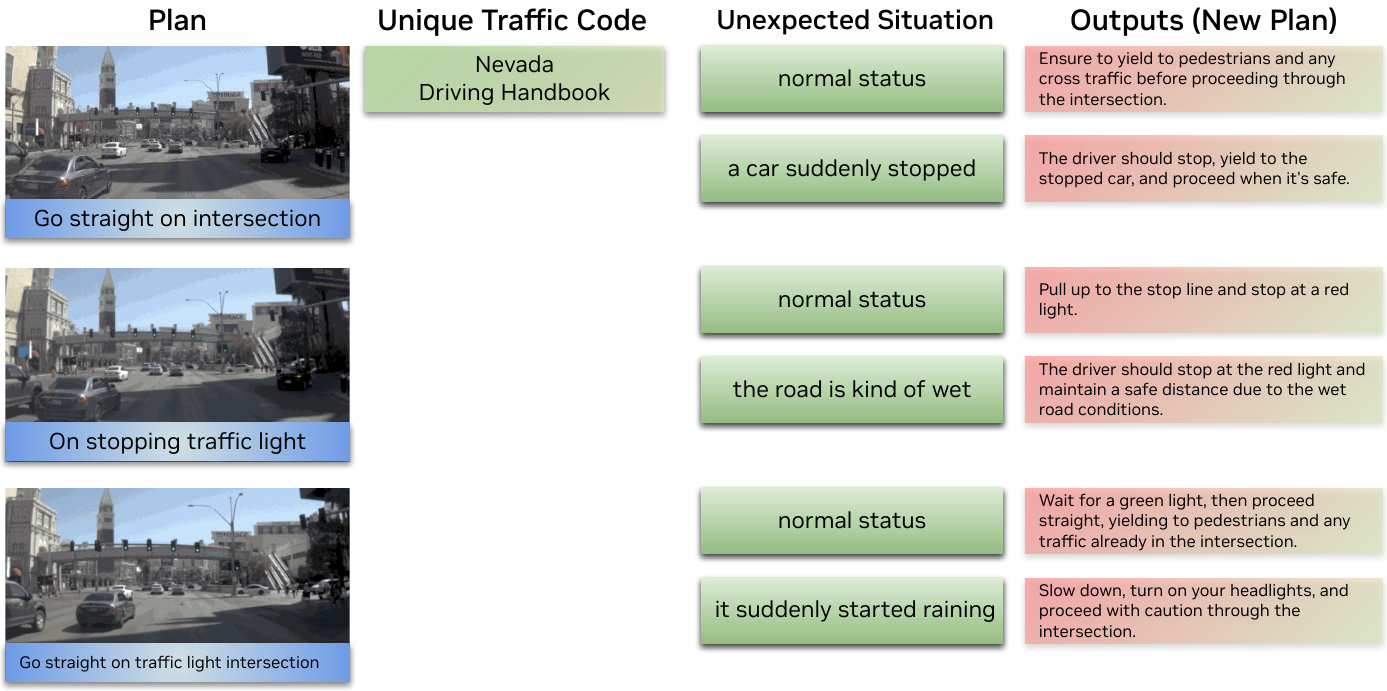}
\caption{Inference on a random Nuplan video in Las Vegas. For this case, we select Nevada driving handbook, the illustration shows that \method{} works very well in Nuplan and can provide accurate new plan with various unexpected situations.}
\label{fig:exp_nuplan}
\end{figure*}

%% file: tables/challenging.tex
\begin{table*}
  \setlength{\tabcolsep}{4pt}
    \centering
    \resizebox{1\linewidth}{!}{
    \begin{tabular}{llllll}\toprule
    Example & Original Location & Target Location& Plan & Unexpected Situation & Outputs (New Plan) \\ 
    \toprule
   1& San Francisco&NYC&Turn right on red&	normal status& Do not turn right on red in NYC unless a sign permitting it is posted.\\
    & NYC&San Francisco&Turn right on red&	normal status& Stop completely, yield for pedestrians and\\
    & &&&	& turn right if there's no "No Turn on Red" sign.\\
    \midrule
    2& California&Germany&Drive straight on the highway&an emergency vehicle is approaching from behind&Move to the right and allow the emergency vehicle to pass.\\
    \midrule
    3& NYC&London&Drive straight on the highway&the car in front drives very slowly, &Overtake the slow car safely from the right lane, \\
    & &&&	we are in the middle lane of a 3 lane highway& as overtaking on the left is illegal in London.\\
    \midrule
    4& California&Singapore&Unprotected right&normal status&  Yield to all other traffic and pedestrians before making your right turn.\\
    \midrule
    5& California&Germany&Drive straight on the highway, in the leftmost lane&I keep getting honked at by cars behind me&Move to the right lane, the leftmost lane in Germany is for overtaking and faster vehicles.\\
    \midrule
    6& California&Ontario&Driving on a rural two-lane road&	there’s a horse pulling a carriage&The driver should slow down, pass the carriage cautiously, and give plenty of space to the horse. \\
    \bottomrule
    \end{tabular}
    }
  \caption{Inference on various challenging examples. We compare the outputs with and without \method{}, and it is obvious that \method{} significantly improves the plan and successfully handles various difficult tasks under diverse scenarios and environments. }
  \label{tab:challenging}
\end{table*}

%% file: imgs/viz/figure.tex
\begin{figure*}[ht]
    \centering
    \begin{minipage}{0.5\textwidth}
        \includegraphics[width=\linewidth]{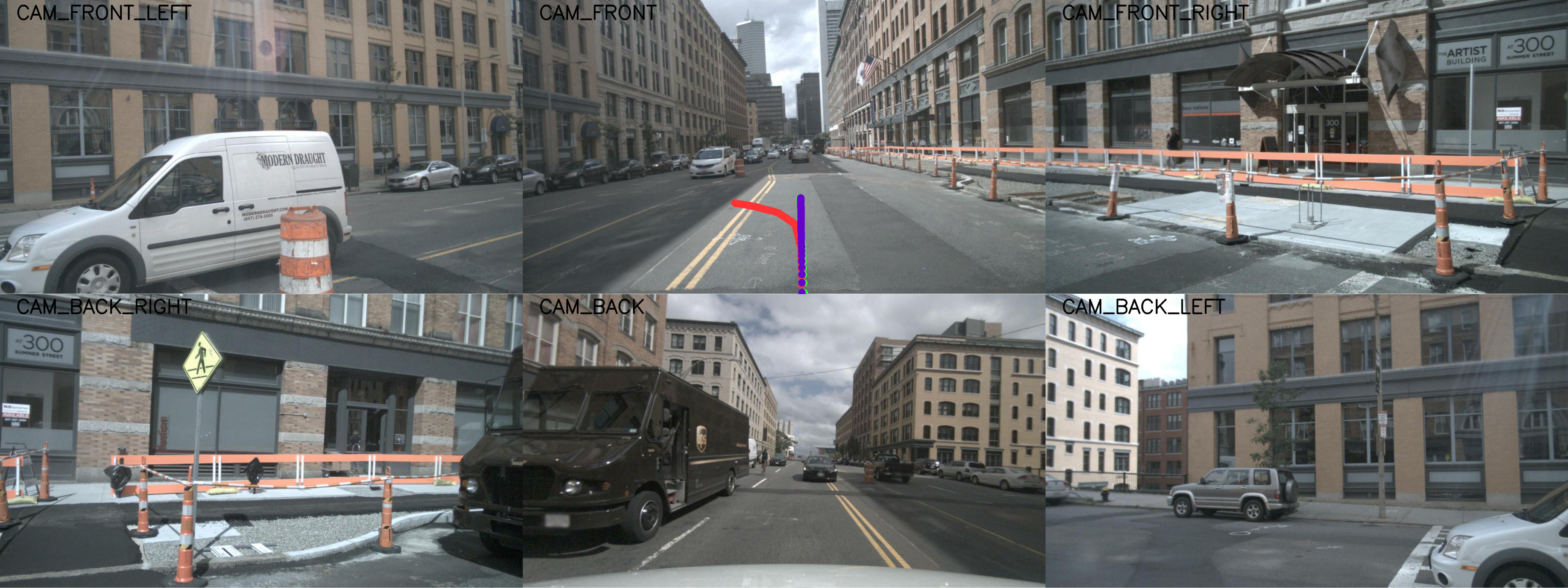}
    \end{minipage}%
    \begin{minipage}{0.5\textwidth}
        \includegraphics[width=\linewidth]{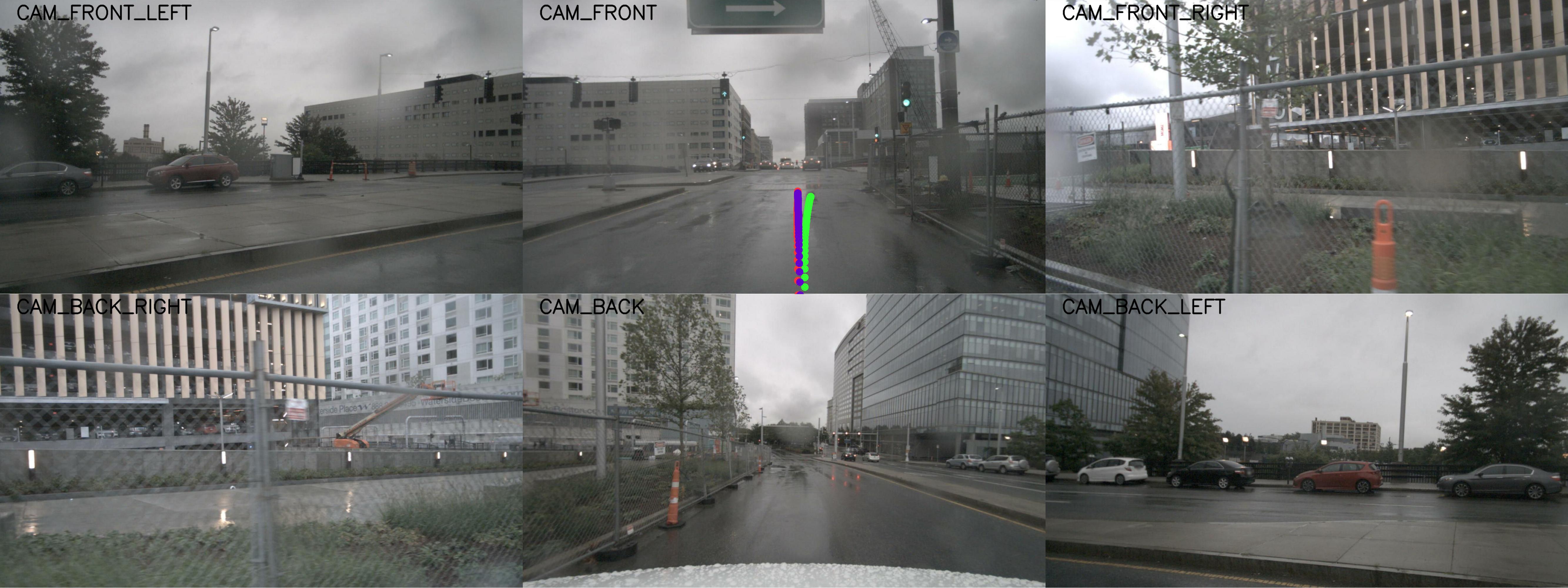}
    \end{minipage}
    \begin{minipage}{0.5\textwidth}
        \includegraphics[width=\linewidth]{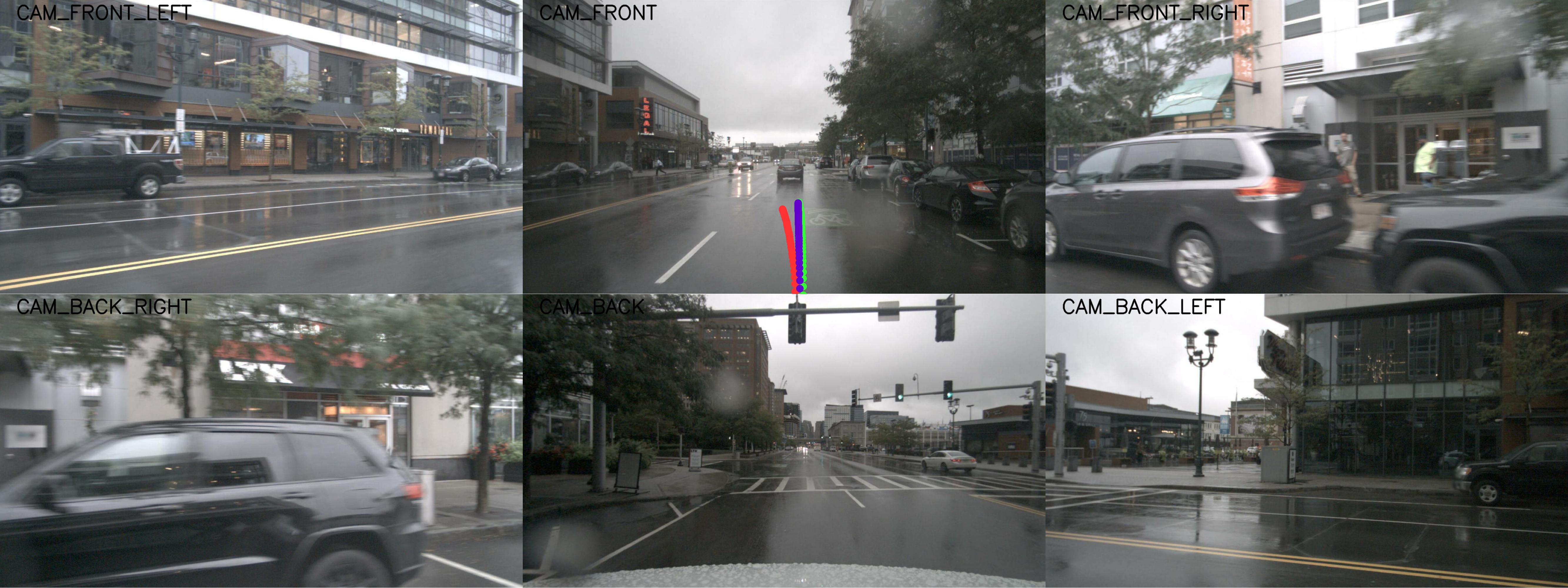}
    \end{minipage}%
    \begin{minipage}{0.5\textwidth}
        \includegraphics[width=\linewidth]{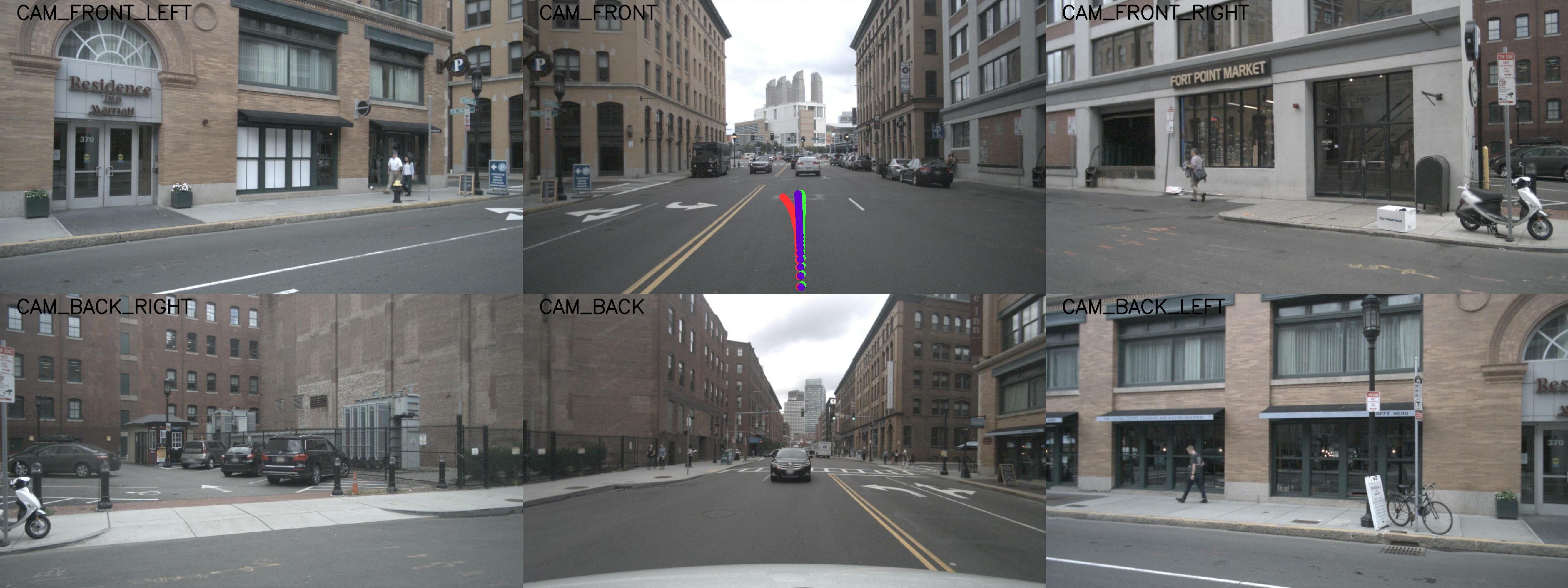}
    \end{minipage}
    \caption{Visualization of the motion planning results on the nuScenes Boston subset. Ground truth trajectories are in \textbf{\greentext{green}}. Trajectories generated by GPT-Driver are in \textbf{\redtext{red}}. Re-planned trajectories by \method{} are in \textbf{\purpletext{purple}}.}
\vspace{-0.1in}
\end{figure*}

%% file: tables/planning.tex
\begin{table*}[]
\begin{center}
\begin{tabular}{l|cccc|cccc}
\toprule
\multirow{2}{*}{Method} &
\multicolumn{4}{c|}{L2 (m) $\downarrow$} & 
\multicolumn{4}{c}{Collision (\%) $\downarrow$} \\
\cmidrule(){2-9}
& 1s & 2s & 3s & \cellcolor{gray!30}Avg. & 1s & 2s & 3s & \cellcolor{gray!30}Avg. \\
\midrule
GPT-Driver~\cite{mao2023gptdriver} & 0.27 & 0.59 & 1.04 & 0.63 & 0.22 & 0.45  & 1.07 & 0.58 \\
\midrule
GPT-Driver~\cite{mao2023gptdriver} + \method{} Re-Planning (Ours) & \textbf{0.27}  & \textbf{0.58} & \textbf{1.02} & \textbf{0.62} & \textbf{0.22} & \textbf{0.41} & \textbf{1.04} & \textbf{0.56}  \\
\bottomrule
\end{tabular} %}
\end{center}
\vspace{-0.2in}
\caption{Motion planning results. With the guidelines provided by \method{}, GPT-Driver could adapt its motion planning capability from Singapore to Boston and reduce planning errors.
}
\label{tab:plan}
\end{table*}

%% file: imgs/exp_youtube/figure.tex
\begin{figure*}
\centering
\includegraphics[width=1.0\linewidth]{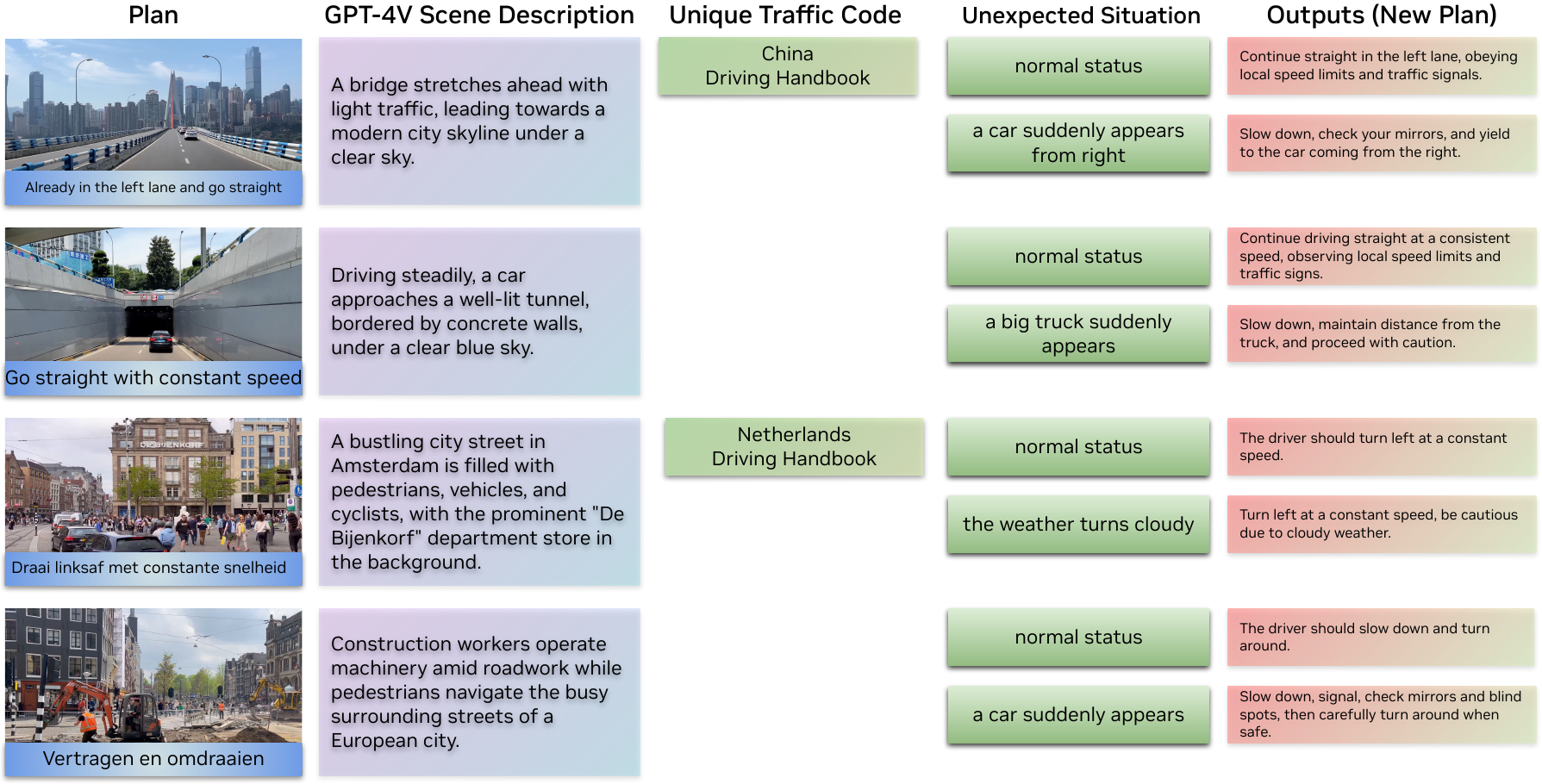}
\caption{Combining \method{} with GPT-4V and inference on random YouTube scenes. In the first two rows, we show the scenes of Chongqing in China. In the rest two rows, we show the scenes of the crowded Amsterdam Center in Netherlands. As for the plans, we input {\color{bleudefrance}{Dutch}} instead of {\color{bleudefrance}{English}} to verify the capability of \method{} to interact with drivers from various cultures automatically without mentioning the language type of the input. ``Draai linksaf met constante snelheid" represents ``Turn left with constant speed" and ``Vertragen en omdraaien" represents ``decelerate and turn around". We asked the system to output English instructions, we noticed that the system extracts keywords in English and is able to provide accurate plans.}
\label{fig:gpt4v}
\end{figure*}

%% file: sec/conclusion.tex
\section{Conclusion, Limitations, and Future Work}
\label{sec:conclusion}
\textbf{Conclusion.}
In this work, we proposed \method{}, an LLM-powered framework that adapts nominal motion plans by a human driver or an AV to local traffic rules of that region. The modularity of \method{} facilitates its use for human driver assistance as well as AV plan adaptation. To our knowledge, \method{} is the first to propose traffic rule-based adaptation via LLMs. Our results show that human drivers find \method{} to be helpful for driving in new locations, and \method{} also improves the performance of AV planners in new locations.

\textbf{Limitations.}
Though \method{} provides various benefits, it also suffers from two limitations: first, since \method{} requires running an LLM in the control loop, the runtime for \method{} is not yet conducive for closed-loop use in an AV planning stack -- this limitation is shared by all LLM-based motion planners. Second, as discussed in our results earlier, \method{} is sensitive to the quality of scene descriptions. Although GPT-4V can provide such descriptions, they are sometimes not sufficiently accurate. This limitation points towards the need to develop an AV-specific foundation model that can provide AV-centric scene descriptions. 

\textbf{Broader Impact.}
As a human driver assistant, we hope that \method{} would reduce the number of road accidents induced by tourists oblivious of the local traffic rules. As a policy adapter for AVs, we expect \method{} to pave the way towards easier traffic rule adaptation for AVs allowing them to expand their operations beyond geo-fenced regions.

\textbf{Future Work.} For future work, there are various directions we are excited to pursue: first, we will explore improving GPT-4V's scene descriptions by fine-tuning it on AV datasets. Second, we will explore the development of an unexpected scenario detector which will allow us to use \method{} only when it is needed, thereby significantly alleviating the computational burden involved in running an LLM-based module in the control loop. Finally, we will work towards furnishing safety certificates for the LLM outputs by leveraging recent developments in uncertainty quantification and calibration techniques for ML, such as conformal prediction and generalization theory.